# Fast Prediction of New Feature Utility

**Hoyt Koepke**                                                                                   HOYTAK@STAT.WASHINGTON.EDU
Department of Statistics University of Washington Seattle, WA 98107

**Mikhail Bilenko**                                                                                      MBILENKO@MICROSOFT.COM
Microsoft Research, Redmond, WA 12345

## Abstract

We study the *new feature utility prediction* problem: statistically testing whether adding a feature to the data representation can improve the accuracy of a current predictor. In many applications, identifying new features is the main pathway for improving performance. However, evaluating every potential feature by re-training the predictor can be costly. The paper describes an efficient, learner-independent technique for estimating new feature utility without re-training based on the current predictor's outputs. The method is obtained by deriving a connection between loss reduction potential and the new feature's correlation with the loss gradient of the current predictor. This leads to a simple yet powerful hypothesis testing procedure, for which we prove consistency. Our theoretical analysis is accompanied by empirical evaluation on standard benchmarks and a large-scale industrial dataset.

## 1. Introduction

In many mature learning applications, training algorithms are advanced and well-tuned, making the discovery and addition of new, informative features the primary driver of error reduction. New feature design strategies include addition of previously unused descriptive signal sources, as well as various methods that derive new features from the existing representation. A newly proposed feature is typically evaluated by augmenting it to the data representation and re-running the training and validation procedures to observe the resulting difference in predictive accuracy. However, re-training carries significant costs in many real-world applications:

- *Computational costs*: large-scale domains (e.g., web search and advertising) often employ computationally expensive learners and large training datasets. This imposes experimentation latency that is a barrier to rapid feature prototyping.
- *Logistical costs*: training processes for industry tasks are often componentized across large infrastructure pipelines, running which requires domain expertise. Potential feature contributors lacking such expertise are hence deterred from evaluating their features by the training pipeline complexity.
- *Monetary costs*: in the domains of medical and finance applications, new feature values may be unavailable for the entire training set or may carry non-negligible costs, calling for methods that predict their utility based on a sampled subset.

These costs call for feature utility prediction methods that do not rely on re-training, instead viewing the learner as a black box constructing a best-possible predictor from a chosen model class. The black-box assumption implies that the only description of the learned predictor is provided via its evaluation on labeled data (e.g., on a hold-out set or via cross-validation), on which its outputs are compared with true target values via a task-appropriate loss function. Thus, our objective is to design a computationally inexpensive algorithm for statistically determining whether adding a new feature can potentially reduce the expected loss, given the current predictor.

To derive a principled algorithm for the problem, we prove that under mild assumptions, testing whether a feature can yield predictive accuracy gains is equivalent to testing its correlation with the negative loss gradient, against which we train a squared-loss regressor. To construct a provably consistent hypothesis test, we form a null distribution by bootstrapping the marginal distributions. The overall algorithm is easily parallelizable, does not require re-training, and works on subsampled datasets, making it particularly appropriate for large data contexts. The method is applicable to a wide variety of learning tasks, requiring only estimates of the functional gradient of the loss, which can be approximated even for discontinuous losses that are common in structured tasks, i.e., ranking.

The rest of the paper is organized as follows: Section 2 describes related work, followed by Section 3 that formally defines the problem and motivates the approach. Section 4 describes our method and provides theoretical analysis. Section 5 summarizes empirical evaluation of the approach, followed by discussion of future work and conclusions in Sections 6 and 7.

## 2. Related Work

The task addressed in this paper – efficient estimation of predictive utility for *new* features without re-training – is related yet distinct from three known problems: feature selection (Guyon & Elisseeff, 2003), active feature acquisition (Saar-Tsechansky et al., 2009), and feature extraction (Krupka et al., 2008). While these tasks also involve estimating measures of feature importance, they have different objectives. Critically, many techniques for these problems rely on re-training, while our motivation is avoiding it.

In contrast to our setting, where the objective is to efficiently triage new features for addition, feature selection aims to remove unnecessary existing features (Guyon & Elisseeff, 2003). Representatives include wrapper approaches that utilize multiple rounds of re-training with feature subsets, methods that use prediction results for instances with permuted or distorted feature values (Breiman, 2001; Kononenko, 1994), and filter techniques that rely on joint statistics of features and class labels (Song et al., 2007).

Feature acquisition aims to incrementally select individual feature values for addition to the dataset via estimating their expected utility, and can be viewed as a feature-focused variant of active learning (Lizotte et al., 2003; Saar-Tsechansky et al., 2009). Proposed solutions rely on expensive value-of-information computation, making them prohibitive for our setting. Feature extraction methods attempt to construct new joint features that combine individual attributes by evaluating their dependency structure (Della Pietra et al., 1997; Krupka et al., 2008). In contrast, our approach seeks to directly evaluate the possibility of improvement in prediction accuracy for new features.

On the theoretical side, several approaches have used bootstrapping or permutation tests to assess predictive value of features (Fromont, 2007; Anderson & Robinson, 2001; Ojala & Garriga, 2010). These methods typically utilize the tests to assess the generalizability of results obtained on the finite sample case, a well-known property (Van der Vaart & Wellner, 1996).

Also of note is recent work on testing for the statistical independence of features, a key component of our analysis (Gretton & Györfi, 2010). In particular, there has been active work for kernel methods that use the Hilbert-Schmidt Independence Criteria (Sriperumbudur et al., 2010; Song et al., 2007; Gretton et al., 2008). While our approach also relies on functional analysis techniques, it provides an alternative that does not rely on kernels, instead using standard correlation methods, similarly in spirit to (Huang, 2010).

## 3. New Feature Utility & Independence

We consider the standard inductive learning setting, where training data is a set of samples of random variable pairs, $(X_i, Y_i)$ from an unknown joint distribution function $P_{X,Y}$, corresponding to data instances described by feature values $X_i$ and prediction targets $Y_i$. Learning corresponds to finding a predictor function $f_0$ from some function class $\mathcal{F}_X$ that minimizes expected loss $\mathbb{E} L(f(X), Y)$ for a given loss function $L$ encoding the application-appropriate error measure:

$$f_0 = \operatorname*{argmin}_{f \in \mathcal{F}_X} \mathbb{E} L(f(X), Y) \quad (1)$$

The new feature utility prediction problem can be posited as determining whether adding an additional random variable, $X'$, to the data representation can result in reduction of expected loss if the predictor was re-trained with it. We designate the function class for predictors on the resulting representation as $\mathcal{F}_{X,X'} = \mathcal{F}$; it subsumes function classes $\mathcal{F}_X$ and $\mathcal{F}_{X'}$ which are restricted to predictors that depend only on feature sets $X$ or $X'$ respectively. Formally,

$$\mathcal{F}_X = \{f \in \mathcal{F} : \exists g \text{ s.t. } f(X, X') \stackrel{\text{a.s.}}{=} g(X)\}$$
$$\mathcal{F}_{X'} = \{f \in \mathcal{F} : \exists g \text{ s.t. } f(X, X') \stackrel{\text{a.s.}}{=} g(X')\}$$

Then, the new feature utility prediction problem can be formalized as the hypothesis test of:

$$(H_1) \quad \min_{f \in \mathcal{F}} \mathbb{E} L(f(X, X'), Y) < \mathbb{E} L(f_0(X), Y)$$

against the null hypothesis $H_0$ in which they are equal. In other words, we define feature utility as the capability of the feature to lower the expected predictor loss in the infinite sample case (i.e., w.r.t. to the true distributions). Thus, we use the theoretical paradigm in which our "test set" is the true distribution.

To motivate the approach, consider the ideal feature evaluation test: determining whether $X' \perp Y \mid X$, i.e. if $X'$ is independent of $Y$ given $X$. If this is answered in the affirmative – the null hypothesis $H_0$ is true – then $X'$ contains no additional information about $Y$ that is not already contained in $X$, and hence the loss cannot be reduced. Otherwise, knowing $X'$ provides information that can be exploited to construct a better predictor as long as $\mathcal{F}$ is sufficiently rich. However, this ideal test is expensive to perform (Huang, 2010; Song, 2009; Su & White, 2008).

Instead of the ideal conditional independence test, we consider a more restricted test that seeks to determine whether $X' \perp \lambda(f_0(X), Y)$ for some function $\lambda$ capable of capturing the part of $Y$ that could not be predicted by $f_0(X)$. If this is answered in the affirmative for an appropriate $\lambda$, then $X'$ contains no additional information about $Y$ that is not already contained in $X$. Otherwise, knowing $X'$ provides some new information that can be utilized to create a better predictor. In the next section, we show that for a broad class of loss functions and predictor classes, we can construct a $\lambda$ and test independence by maximizing the correlation between $g(X')$ and $\lambda(f_0(X), Y)$ for $g \in \mathcal{F}_{X'}$. We show this test to be equivalent to testing:

$(H_1')$ $\exists g \in \mathcal{F}_{X'}$ such that
$$\mathbb{E} L(f_0(X) + g(X'), Y) < \mathbb{E} L(f_0(X), Y)$$

against the null $(H_0')$ in which such $g$ does not exist. In the following sections, we develop a consistent test for $(H_1')$ against $(H_0')$ under mild regularity assumptions on the predictor class and the loss function $L(\cdot, Y)$.

## 4. A Consistent Feature Utility Test

In this section, we present a theoretical description of our approach and prove, under reasonable assumptions, that it provides an accurate test of whether a new feature $X'$ can improve prediction performance. The key part of the proof, detailed in Section 4.2, is the use of the bootstrap to test for the statistical independence of the new feature to a residual function of the current predictions. To set up this test, we first list and discuss several assumptions on the predictor class and the loss function. Then, in Section 4.1, we show a sequence of equivalent formulations of our problem in the context of the true joint distribution of $(X, X', Y)$. The goal, reached in statement T4 of Theorem 4.2, is a formulation that can be accurately tested in the finite sample case using the bootstrap. This formulation, combined with a way to handle the optimization component of the bootstrap test, leads to the practical algorithm presented in Section 4.3.

**Loss Function Assumptions.** Our assumptions on the loss minimized when searching for a predictor from $\mathcal{F}$ are quite weak: finiteness, a type of strict monotonicity, and an available direction of descent:

**L1.** Finiteness: $\forall f \in \mathcal{F}$, $\mathbb{E} |L(f(X, X'), Y)| < \infty$.
**L2.** Weak augmenting functional convexity:
  For all $f \in \mathcal{F}_X$ and $g \in \mathcal{F}_{X'}$ such that $\mathbb{E} L(f(X), Y) \geq \mathbb{E} L(f(X) + g(X'), Y) + \eta$ for some $\eta > 0$, there $\exists \beta > 0$ s.t. $\forall \alpha \in [0, 1]$, $\mathbb{E} L(f(X), Y) \geq \mathbb{E} L(f(X) + \alpha g(X'), Y) + \alpha \beta \eta$.
**L3.** A functional descent direction:
  Let $f_0$ be optimal in $\mathcal{F}_X$ as given by (1). Then either $f_0$ is also optimal in $\mathcal{F}$, or there exists a random variable $\Lambda_{f_0}$ dependent on $(X, X', Y)$, with $\text{std}(\Lambda_{f_0}) = 1$, such that, $\forall h \in \mathcal{F}$ in a sufficiently small neighborhood of $f_0$, $\mathbb{E}[\Lambda_{f_0}(h(X, X') - f_0(X))] > 0$ implies that $\mathbb{E} L(h(X, X'), Y) < \mathbb{E} L(f(X), Y)$.

Condition (L2) essentially imposes a type of strict monotonicity on the function class away from the minimum. It is much weaker than convexity; all it requires is that moving in the direction of a better optimizer gives *some* improvement, even if it is relatively small. Condition (L3) intuitively says there exists a direction along which improvement in the expected loss is guaranteed – provided improvement is possible.

These assumptions are quite weak and cover many non-convex, discontinuous loss functions. In the case of a convex, differentiable loss function, however, we show below that conditions (L2) and (L3) are satisfied and $\Lambda_{f_0}$ has an easy and natural form.

**Prediction Functions Assumptions.** The assumptions on the classes of prediction functions $\mathcal{F}$, $\mathcal{F}_X$ and $\mathcal{F}_{X'}$ needed to prove consistency are the following:

**F1.** Closure under scaling: $cf \in \mathcal{F}$ $\forall f \in \mathcal{F}, c \in \mathbb{R}^+$.
**F2.** Closure under shifting: $d + f \in \mathcal{F}$ $\forall f \in \mathcal{F}, d \in \mathbb{R}$.
**F3.** $\min_{f \in \mathcal{F}} \mathbb{E} L(f(X, X'), Y) \leq \min_{g \in \mathcal{F}_{X'}} \mathbb{E} L(f_0(X) + g(X'), Y)$.
**F4.** $\forall f \in \mathcal{F}$, $f(X, X')$ is bounded, or, generally, $\mathcal{F}$ is $P$-Donsker (Van der Vaart & Wellner, 1996).

These conditions, while seemingly obscure, are generally satisfied by most modern predictor classes. Condition (F3) states that training on all features will result in a better predictor than one obtained by first training on a subset of features, then "patching" it with the remaining features. This bound guarantees that the improvement in loss on the full predictor can only be greater than that obtained by a decomposed model.

The assumption (F4) – that $\mathcal{F}$ is $P$-Donsker – bounds the flexibility of the class of classifiers. Intuitively, it means that when working with an asymptotically large sample, the behavior of the classifier is not inordinately dominated by a few outlier values. This assumption ensures the behavior of the bootstrap is reasonable – two classifiers trained on different bootstrapped samples should not be wildly different. The boundedness of $f \in \mathcal{F}$, along with measurability assumptions, implies this (Van der Vaart & Wellner, 1996), but virtually all machine learning algorithms used in practice satisfy this assumption.

**Correlated Features and the XOR Problem.** An obvious question to ask is: what about new features that are only useful in conjunction with existing features (of which $Y = X_1$ XOR $X_2$, with $X_1, X_2 \overset{\text{iid}}{\sim} \mathcal{B}ernoulli(\tfrac{1}{2})$ is the canonical example), wouldn't assumption (L3) be violated and the approach fail to predict their utility? This intuition is correct; however, when this situation occurs in practice, it is rarely in the absence of other modeling information indicating that it might be the case. As our method allows for multiple variables to be tested as a block, or existing features to be recycled by setting one or more dimension of $X'$ to specific dimensions of $X$, such cases can easily be handled.

**Connection to Convex Loss Functions.** In the case of a convex, differentiable loss functions, we show that (L2) and (L3) are satisfied, and direction of descent $\Lambda_{f_0}$ can be defined by the distribution of the negative gradient of the loss function, making it easily computable in practice.

**Theorem 4.1:** *Suppose that for $\forall y$ in the support of $Y$, $L(u, y)$ is convex and differentiable in $u$ and satisfies assumption (L1). Then (L2) and (L3) hold with $\Lambda_{f_0}$ given by*

$$\Lambda_{f_0} = -\frac{1}{\sigma}\lambda(f_0(X), Y), \quad \lambda(u,y) = \frac{\partial}{\partial u}L(u,y) \quad (2)$$

*where $\sigma$ is defined to produce $\operatorname{std}(\Lambda_{f_0}) = 1$.*

*Proof.* (L2) immediately follows; $\forall f \in \mathcal{F}_X$ and $h \in \mathcal{F}_{X'}$,

$$L(f(X), Y) - L(f(X) + \alpha h(X'), Y)$$
$$\leq \alpha \left[ L(f(X), Y) - L(f(X) + h(X'), Y) \right]$$

by the definition of convexity; taking expectations yields the result. Now, using (L1) and the differentiability of $L$, it is easy to show that the Gateaux functional derivative $d\Gamma(f; h)$ (Van der Vaart, 2000) of the functional $\Gamma(f) = \mathbb{E} L(f(X, X'), Y)$ is given by $d\Gamma(f; h) = \mathbb{E}[\lambda(f(X, X'), Y) h(X, X', Y)]$, with $\lambda$ given in (2). Now, $d\Gamma(f; h)$ defines a linear operator in functional space in which $d\Gamma(f; h)$ gives the change in $\Gamma(f)$ in the direction $h$. (L3) and (2) immediately follow from geometry. □

### 4.1. Equivalent Tests

In this section, we show that testing $(H_1')$ is equivalent to testing for the existence of a feature transform positively correlated with the loss gradient, which allows designing a consistent bootstrap algorithm for it.

When evaluating a new feature, we are interested in finding a function in $\mathcal{F}_{X'}$ that improves the expected loss. In light of condition (L3), define

$$g_0 = \operatorname*{argmax}_{g \in \mathcal{F}_{X'} \,:\, \operatorname{std}(g(X'))=1} \mathbb{E}\, g(X') \Lambda_{f_0}. \quad (3)$$

as the function that most closely aligns with a direction of improvement. The following theorem connects improvement in expected loss to this function.

**Theorem 4.2:** *Suppose the loss function $L$ and predictor class $\mathcal{F}$ satisfy conditions (L1)-(L3) and (F1)-(F4). Let $f_0 = \operatorname{argmin}_{f \in \mathcal{F}_X} \mathbb{E} L(f(X), Y)$, and let $\Lambda_{f_0}$ and $g_0$ be as defined in (L3) and Eq. (3). Then the following are equivalent:*

*T1.* $\exists g \in \mathcal{F}_{X'}$ *that improves the expected loss:*
 $\mathbb{E} L(f_0(X) + g(X'), Y) < \mathbb{E} L(f_0(X), Y).$
*T2.* $\min_{\beta \in \mathbb{R}^+} \mathbb{E} L(f_0(X) + \beta g_0(X'), Y) < \mathbb{E} L(f_0(X), Y).$
*T3.* $\mathbb{E}\left[g_0(X') \cdot \Lambda_{f_0}\right] > 0.$
*T4.* $\mathbb{E}\left[g_0(X') \cdot \Lambda_{f_0}\right] - \mathbb{E} g_0(X') \mathbb{E} \Lambda_{f_0} > 0.$

*Proof.* First, we show that T1 implies T3. Suppose that T1 is true. By (L2), $\mathbb{E} L(f_0(X) + \alpha g(X'), Y) < \mathbb{E} L(f_0(X), Y)$ for all $\alpha \in (0, 1]$. As $\alpha g(X') \in \mathcal{F}_{X'} \ \forall \alpha$, by (F1), $\mathbb{E}\left[(\alpha g(X')) \cdot \Lambda_{f_0}\right] > 0$ for $\alpha > 0$ sufficiently small. Now, by (F1), $g(X')/\operatorname{std} g(X') \in \mathcal{F}_{X'}$, thus

$\mathbb{E}\left[g_0(X') \cdot \Lambda_{f_0}\right] \geq \mathbb{E}\left\{[g(X')/\operatorname{std}(g(X'))] \cdot \Lambda_{f_0}\right\} > 0$

as $g(X')/\operatorname{std}(g(X'))$ is included in the optimization of equation (3). Now, by (L3), T3 implies that $\mathbb{E} L(f_0(X) + \beta_0 g_0(X'), Y) < \mathbb{E} L(f_0(X), Y)$ for some $\beta_0 > 0$ sufficiently small. T2 follows, as

$\min_{\beta \in \mathbb{R}^+} \mathbb{E} L(f_0(X) + \beta g_0(X'), Y)$
$\leq \mathbb{E} L(f_0(X) + \beta_0 g_0(X'), Y) < \mathbb{E} L(f_0(X), Y)$

T2 trivially implies T1, thus T1 - T3 are equivalent. For the equivalence of T3 and T4 we show that, given the closure of $\mathcal{F}$ under constant shifts, $\mathbb{E} \Lambda_{f_0} = 0$.

Suppose $\mathbb{E} \Lambda_{f_0} = c \neq 0$. Let $\varepsilon > 0$, and consider the function $g_0(\cdot) = \varepsilon c$. Now $\mathbb{E} \varepsilon c \Lambda_{f_0} = \varepsilon c^2 > 0$. As $f_0(\cdot) + \varepsilon c \in \mathcal{F}_X \subset \mathcal{F}$ by (F2), (L3) implies that, for $\varepsilon$ sufficiently small, $\mathbb{E} L(f_0(X) + \varepsilon c, Y) < \mathbb{E} L(f_0(X), Y)$. This contradicts the optimality of $f_0$ in $\mathcal{F}_X$; thus $\mathbb{E} \Lambda_{f_0} = 0$, making T3 equivalent to T4. □

### 4.2. A Consistent Hypothesis Test

The above proofs work for random variables with respect to their true distributions; the bridge between this and a practical algorithm is the bootstrap. As discussed earlier, we are interested in assessing the performance of our predictor on the true distribution, which requires a consistent test of whether

$(H_1^*)\ \exists\, g \in \mathcal{F}_{X'}$ s.t. $\mathbb{E}\left[g(X') - \mathbb{E} g(X')\right]\left[\Lambda_{f_0} - \mathbb{E}\Lambda_{f_0}\right] > 0$

against the null, where equality holds. By Theorem 4.2, we have that test $(H_1^*)$ is equivalent to $(H_1')$.

In the next theorem, we define an accurate hypothesis test of $(H_1^*)$ and prove its consistency. The last step needed is a good estimator $\hat{\Lambda}_{f_0}$ of $\Lambda_{f_0}$ generated by the minimum-risk predictor $f_0 \in \mathcal{F}_X$ on the true distribution. This is because we show the bias in the bootstrap to be controlled by the standard deviation of $\sqrt{n}(\Lambda_{f_0} - \hat{\Lambda}_{f_0})$; in general, this is non-zero.

There are several ways to effectively control this bias. One can assume that the true $\Lambda_{f_0}$ is known or comes from training on a much larger dataset, against which the performance is actually evaluated. This is a com-mon scenario for many large-data domains. Also, one can use methods known to asymptotically reduce the bias, such as k-fold cross validation (Cornec, 2010).

**Theorem 4.3:** *Let*
$$K(U_n, V_n) = \sqrt{n}\left\{\max_{g \in \mathcal{F}_{X'}}\left(\mathbb{E}\,g(U_n)V_n - [\mathbb{E}\,g(U_n)][\mathbb{E}\,V_n]\right)\right\}$$
*Let $\tilde{X}' \stackrel{D}{=} X'$ be a bootstrap sample of $X'$, and let $\tilde{\Lambda}_{f_0} \stackrel{D}{=} \hat{\Lambda}_{f_0}$ be a bootstrap sample of $\hat{\Lambda}_{f_0}$, independent from $\tilde{X}'$. Let $F$ be the c.d.f. of $K(\tilde{X}', \tilde{\Lambda}_{f_0})$, with quantile function $F^{-1}(t) = \inf\left\{u > 0 : \mathbb{P}(K(\tilde{X}', \tilde{\Lambda}_{f_0}) > u) \leq t\right\}$. Fix $\alpha \in (0,1)$, and set the critical point $c_\alpha = F^{-1}(\alpha)$. The test that accepts the alternative hypothesis, $(H_1^*)$, for values of $K(X', \Lambda_{f_0}) \geq c_\alpha$, and rejects otherwise, has asymptotic level $\alpha$ and bias at most $\max(F(c_\alpha + \eta) - \alpha,\ \alpha - F(c_\alpha - \eta))$, where $\eta = \sqrt{8n}\sqrt{1 - \mathbb{E}_{P_n}\Lambda_{f_0}\hat{\Lambda}_{f_0}}$.*

*Proof.* Drop the subscript $f_0$ from $\Lambda$ and $\hat{\Lambda}$ for convenience. From Th.4.2, we need a consistent test of
$$\|H - P \times Q\|_\mathcal{G} = \max_{g \in \mathcal{F}_{X'}} \left|\mathbb{E}\,g(X')\hat{\Lambda} - \mathbb{E}\,g(X')\,\mathbb{E}\,\hat{\Lambda}\right| > 0$$
where $H$ is the joint measure of $X'$ and $\hat{\Lambda}$, $P, Q$ are the respective marginal distributions, and $\mathcal{G} = \mathcal{F}_{X'} \times \{\mathbf{I}\}$, where $\mathbf{I}$ is the identity function. However, only their empirical samples $H_n$, $P_n$, and $Q_n$ are available.

Let $\hat{P}_n$ and $\hat{Q}_n$ be the measures formed from an independent bootstrap sample of $X'_n = (x'_1, ..., x'_n)$ and $\hat{\Lambda} = (\hat{\lambda}_1, ..., \hat{\lambda}_n)$, and let $\hat{H}_n$ be the joint measure formed from $\hat{P}_n$ and $\hat{Q}_n$. Then let
$$\hat{Z}_n = \sqrt{n}\left\|\hat{H}_n - \hat{P}_n \times \hat{Q}_n\right\|_\mathcal{G},\ c_\alpha = \inf\left\{c : P(\hat{Z}_n > c) \leq \alpha\right\}$$
under these conditions, we reject the null if
$$\sqrt{n}\,\|H_n - P_n \times Q_n\|_\mathcal{G} > c_\alpha.$$
From (Van der Vaart & Wellner, 1996), this test is consistent with asymptotic level $\alpha$. To complete the proof, note that
$$\max_{g \in \mathcal{F}_{X'}} \left|\mathbb{E}\,g(X')\hat{\Lambda} - \mathbb{E}\,g(X')\,\mathbb{E}\,\hat{\Lambda}\right|$$
$$= \max_{g \in \mathcal{F}_{X'}} \left|\mathbb{E}\,g(X')\Lambda - \mathbb{E}\,g(X')\,\mathbb{E}\,\Lambda\right|$$
$$\oplus \max_{g \in \mathcal{F}_{X'}} \left|\mathbb{E}\,g(X')(\Lambda - \hat{\Lambda}) - \mathbb{E}\,g(X')\,\mathbb{E}(\Lambda - \hat{\Lambda})\right|$$
where $a = b \oplus c$ denotes $|a - b| \leq c$. Now
$$\max_{g \in \mathcal{F}_{X'}} \left|\mathbb{E}\,g(X')(\Lambda - \hat{\Lambda}) - \mathbb{E}\,g(X')\,\mathbb{E}(\Lambda - \hat{\Lambda})\right|$$
$$\leq \max_{g \in \mathcal{F}_{X'}} \left|\mathbb{E}\,g(X')(\Lambda - \hat{\Lambda})\right| + \left|\mathbb{E}\,g(X')\,\mathbb{E}(\Lambda - \hat{\Lambda})\right|$$
$$\leq \max_{g \in \mathcal{F}_{X'}} 2\sqrt{\mathbb{E}\,g^2(X')\,\mathbb{E}(\Lambda - \hat{\Lambda})^2} \quad (4)$$
$$\leq 2\sqrt{\mathbb{E}(\Lambda^2 - 2\Lambda\hat{\Lambda} + \hat{\Lambda}^2)} = 2\sqrt{2}\sqrt{1 - \mathbb{E}\,\Lambda\hat{\Lambda}} \quad (5)$$
where (4) follows from Liapunov's and Jensen's inequalities, and (5) uses $\text{std}(\Lambda) = 1$. Combing this result with the bootstrap completes the proof. □

### 4.3. A Feature Evaluation Algorithm

Performing the test requires obtaining a transform that maximizes the inner product between the function and the negative loss gradient. The following theorem allows doing this by via squared-error loss minimization for an appropriately weighted expectation.

**Theorem 4.4:** *Suppose $\mathcal{F}$ satisfies assumptions (F1) and (F2), and let $\Lambda_{f_0}$ be defined as in (L3). Let*
$$f_1^* = \operatorname*{argmin}_{g \in \mathcal{F}_{X'}} \mathbb{E}\left[g(X') - (\Lambda_{f_0} - \mathbb{E}\,\Lambda_{f_0})\right]^2$$
$$f_2^* = \operatorname*{argmax}_{g \in \mathcal{F}_{X'}\,:\,\text{std}(g(X'))=1} \mathbb{E}\left[g(X')\Lambda_{f_0}\right] - \left[\mathbb{E}\,g(X')\right]\left[\mathbb{E}\,\Lambda_{f_0}\right]. \quad (6)$$
*Then $f_2^*(\cdot) \stackrel{a.s.}{=} f_1^*(\cdot)/\text{std}(f_1^*(X'))$.*

*Proof.* Let $Z = (\Lambda_{f_0} - \mathbb{E}\,\Lambda_{f_0})$. Now, consider Eq.(6). We can enforce $\text{std}(g(X')) = 1$ as follows:
$$f_2^* = \operatorname*{argmax}_{g \in \mathcal{F}_{X'}} \mathbb{E}\left\{\left[\frac{g(X')}{\text{std}(g(X'))}\right]Z\right\} = \operatorname*{argmax}_{g \in \mathcal{F}_{X'}} \frac{\mathbb{E}(g(X')Z)}{\text{std}(g(X'))}.$$
This is invariant to scaling of $g$, and $\mathcal{F}_{X'}$ is closed under scaling, so $g$ is scaled to make $\text{std}(g(X')) = \text{std}(f_1^*(X'))$. Furthermore, $\mathbb{E}\{(g(X')+c)Z\} = \mathbb{E}\,g(X')Z + \mathbb{E}\,cZ = \mathbb{E}\,g(X')Z$, so $g$ is invariant to shifts. As $\mathcal{F}_{X'}$ is closed under shifts, $g$ is shifted so that $\mathbb{E}\,g(X') = 0$. Thus
$$g^* = \operatorname*{argmax}_{g \in \mathcal{G}} \mathbb{E}\,g(X')Z = \operatorname*{argmin}_{g \in \mathcal{G}} \mathbb{E}\,g^2(X') - 2\,\mathbb{E}\,g(X')Z + \mathbb{E}\,Z^2$$
where
$$\mathcal{G} = \{g \in \mathcal{F}_{X'} : \text{std}(g(X')) = \text{std}(f_1^*(X')),\ \mathbb{E}\,g(X') = 0\}.$$
Thus $g^*$ is exactly $f_1^*$, proving the theorem. □

The practical implication of the theorem is that for new feature values $\mathbf{X}'_n = (X'_1, ..., X'_n)$ and standardized gradient samples $\hat{\mathbf{\Lambda}}_n = (\hat{\Lambda}_1, ..., \hat{\Lambda}_n)$, $K$ in Theorem 4.3 becomes
$$g^* = \operatorname*{argmin}_{g \in \mathcal{F}_{X'}} \frac{1}{n}\sum_{i=1}^n \left[g(X'_i) - \hat{\Lambda}_i\right]^2 = \operatorname*{argmin}_{g \in \mathcal{F}_{X'}} \mathbb{E}\left[g(\mathbf{X}'_n) - \hat{\mathbf{\Lambda}}_n\right]^2$$
$$K(\mathbf{X}'_n, \hat{\mathbf{\Lambda}}_n) = \sqrt{n}\left[\mathbb{E}\,g^*(\mathbf{X}'_n)\hat{\mathbf{\Lambda}}_n - \mathbb{E}\,g^*(\mathbf{X}'_n)\cdot\mathbb{E}\,\hat{\mathbf{\Lambda}}_n\right] \quad (7)$$
which corresponds to least-squares regression *regardless of loss $L$*. This surprising result allows reducing the new feature utility problem for a wide array of learning tasks and loss functions to a standard task for which powerful algorithms are readily accessible.

Using the bootstrap, this method is turned into a rigorous test for feature significance, summarized in Algorithm 1. The *p*-value score corresponds to rejecting or accepting the hypothesis that the new value will lead to loss reduction. The algorithm also outputs the number of null standard deviations by which the test statistic $v$ is above the null mean (the z-score), here refered to as the utility score. As empirical evaluation demonstrates, this score provides an accurate measure

**Algorithm 1**: Feature Relevance Test

**Input**: $(X'_i, \hat{\Lambda}_i)$, $i = 1, ..., n$.
**Output**: Relevance Score of $X'$ ($p$-value).
$v \leftarrow K(\mathbf{X}'_n, \hat{\mathbf{\Lambda}}_n)$,    // $K$ defined in (7).
**for** $i = 1, ..., N_{bootstrap}$ **do**
  $\tilde{\mathbf{X}}' \leftarrow$ i.i.d. sample of $n$ values from $\mathbf{X}'_n$.
  $\tilde{\mathbf{\Lambda}}^\dagger \leftarrow$ i.i.d. sample of $n$ values from $\hat{\mathbf{\Lambda}}_n$.
  $\tilde{\mathbf{\Lambda}} \leftarrow (\tilde{\mathbf{\Lambda}}^\dagger - \text{mean}(\tilde{\mathbf{\Lambda}}^\dagger))/\text{std}(\tilde{\mathbf{\Lambda}}^\dagger)$
  $t_i \leftarrow K(\tilde{\mathbf{X}}', \tilde{\mathbf{\Lambda}})$,
**return** *utility score as* $(v - \text{mean}(\mathbf{t}))/\text{std}(\mathbf{t})$,
    *p-value as prop. of $t_1,...,t_n$ greater than $v$.*

of relative feature utility, allowing to rank features for which no null statistic $t_i$ is greater than $v$.

The method scales well to large-data tasks as the $N_{\text{bootstrap}} + 1$ evaluations of $K$ can be easily parallelized, $X'$ is typically lower-dimensional than $X$, and efficient distributed algorithms for least-squares regression are well-studied (Bekkerman et al., 2012).

It is important to note that training a regressor $g$ to maximize correlation with the loss function gradient is central to AnyBoost and MART views of boosting as gradient descent in function space (Mason et al., 1999; Friedman, 2001). Analogously, our approach can be viewed as coordinate descent in function space.

## 5. Experimental Evaluation
### 5.1. Datasets

We evaluate the proposed approach on three learning tasks: calibrated binary classification, regression and ranking. Standard loss functions are used for each task: cross-entropy (log-loss) for calibrated classification, squared loss for regression, and NDCG for ranking (Järvelin & Kekäläinen, 2002). Despite the fact that NDCG is discontinuous, it satisfies assumptions (L1)-(L3), and its pointwise functional gradient estimates can be approximated by aggregating pairwise cost differentials as described in (Burges, 2010).

For classification and regression, we use standard real-task benchmarks from the UCI collection, ADULT and HOUSING. For ranking, we employ a large-scale industrial search engine dataset, WEBRANKING. While it uses thousands of individual features, they are grouped into several dozen distinct information sources. Each information source captures some document property and yields multiple numeric features derived from the property for a given query. For example, the *DocumentBody* source yields features based on the document's text contents (e.g., various similarity measures w.r.t. the query), while the *DocumentAnchorText* source yields analogous features based on the annotations of the document's incoming links. The operational setting for feature utility prediction in this domain is to triage potential new information sources considered for addition to the index, reducing the computational and logistical costs that full re-training would involve. Hence, we overload terminology and refer to each source as a multi-dimensional "feature".

Table 1 summarizes the datasets and loss functions used in the experiments. We employ 10-fold cross-validation for experiments on ADULT and HOUSING, and hence the number of instances refers to the entire dataset size. For WEBSEARCH, the number of instances refers to the size of the validation fold (the training set is much larger), and the number of features refers to the number of information sources evaluated.

### 5.2. Methodology

Accuracy of feature utility prediction is evaluated w.r.t. actual error improvements obtained via retraining with the new feature included. Experimental procedure can be summarized as follows:

1. Given dataset $X$ comprised of $d$ features, $X^{(1)}..X^{(d)}$, perform evaluation with all features included to obtain complete data loss $L(f^*)$.
2. For each feature $X^{(i)}$, perform evaluation on an ablated dataset which excludes the feature, obtaining $d$ corresponding predictors $f_0^{(\neg i)}, i = 1..d$. Difference in accuracy $\Delta L_i = L(f^*) - L(f_0^{(\neg i)})$ for each predictor defines the actual utility of feature $X^{(i)}$. For each trained predictor, per-instance values of loss gradient $\hat{\Lambda}^{\neg i}$ are obtained.
3. Using Algorithm 1, a $p$-value and utility score for each held-out feature is computed, where the utility score is the correlation $K(X^{(i)}, \hat{\Lambda}^{\neg i})$ normalized w.r.t. bootstrap-based null distribution.
4. Feature scores and $p$-values are compared to the actual utilities $\Delta L_i$.

*Table 1.* Dataset summary.

| Name | Task | #Features | #Instances | Loss | Loss gradient, $\frac{\partial}{\partial f}L(f, y)$ |
|---|---|---|---|---|---|
| ADULT | Classification | 14 | 45222 | Cross-entropy | $\frac{f-y}{f(f-1)}$ |
| HOUSING | Regression | 13 | 506 | Squared loss | $f - y$ |
| WEBSEARCH | Ranking | 26 | 741,325 | NDCG | $\lambda$-estimates (Burges, 2010) |

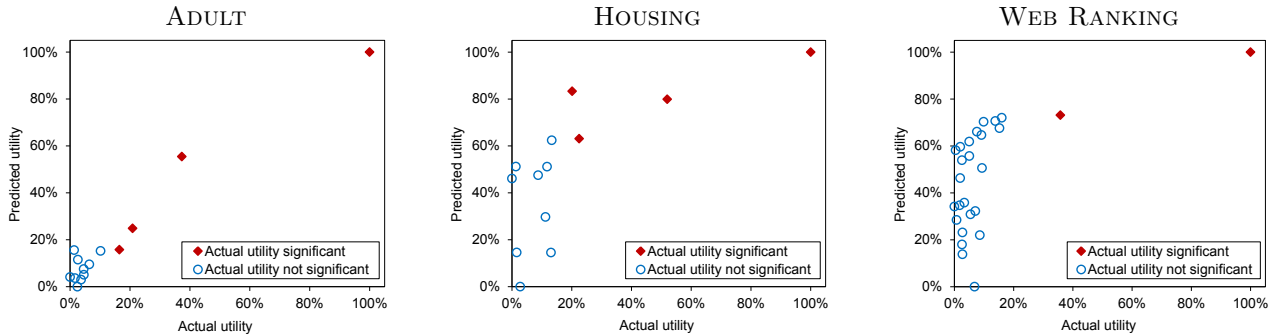

Figure 1. Predicted vs. actual feature utility

In the above procedure, "evaluation" in steps 1 and 2 refers to 10-fold cross-validation for UCI datasets, and training followed by testing on the validation fold for WEBSEARCH. Gradient boosted trees were used for all tasks (Friedman, 2001), using training loss corresponding to each task. For ranking, the LambdaMART tree boosting algorithm that optimizes NDCG was used (Burges, 2010). Solving for optimal $g(\cdot)$ with maximal correlation to negative loss gradient and the corresponding bootstrap trials were performed using boosted regression trees, optimizing for squared-error loss as dictated by Theorem 4.4. Bootstrapping was performed for 100 rounds.

### 5.3. Results and Discussion

Per-dataset plots in Figure 1 illustrate predicted vs. actual utilities for each feature, reported as percentages of the range obtained across all features, with actual utilities based on loss reduction due to feature being added, and predicted utilities based on scores produced by Algorithm 1. In other words, the feature with highest actual and predicted utility appears at 100% on horizontal and vertical axes, respectively. Features for which actual utility is significant at $p < 0.05$ (over validation folds) are demarkated.

As the results demonstrate, the proposed method identifies the features that produce actual accuracy gains with very high recall: all features that are determined to be insignificant indeed produce no meaningful accuracy gains. While some of the features identified as relevant did not in fact produce sizable accuracy gains, this is expected: while a feature may have some predictive value, the predictor class or learning algorithm may be unable to realize it. The practical motivation for the problem is feature triage, where a feature engineer seeks to quickly prioritize features by their potential for improving prediction quality, and the results demonstrate that our approach indeed provides such prioritization accurately.

**Comparison with Feature Selection Heuristics.** We also evaluated several commonly used feature selection heuristics that do not rely on re-training. Figure 2 illustrates the performance of $\chi^2$ Statistic, Information Gain Ratio, and Correlation-based Feature Selection (CFS) (Guyon & Elisseeff, 2003; Hall, 1999) for the ADULT dataset. These results demonstrate that methods that compute feature utility greedily ($\chi^2$ and Information Gain Ratio) can significantly overestimate the value of features that are not informative given others, as evidenced by the two top-scoring features that have near-zero actual utility (in top left corner of corresponding figures). CFS works better as it takes into account the new feature's correlation with other features as well as the label, yet it underestimates the utility of the best feature dramatically, demonstrating the shortcoming of label-based estimates vs. utilizing losses of the current predictor used by our approach.

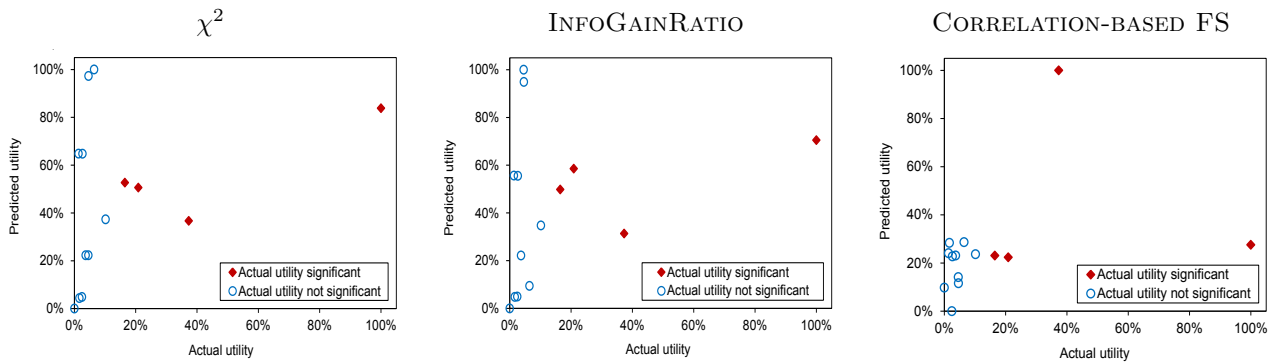

Figure 2. Performance of feature selection heuristics (ADULT dataset)

## 6. Future Work

While this paper demonstrated that the feature utility prediction problem can be solved by posing it as a hypothesis test in function space, it would be interesting to see alternative algorithms for the problem designed via information-theoretic formulations. Another potentially fruitful direction for future work is developing semi-supervised methods that can utilize unlabeled data for improving new feature utility estimates, given its abundance in large-scale domains. Additionally, designing modifications of the described approach for feature selection, extraction and active feature-value acquisition could yield new efficient methods for these tasks, as the overall idea of exploiting outputs of an existing predictor is clearly relevant for these problems. Finally, another attractive future work direction lies along creating new feature utility prediction algorithms that remove the "black-box" assumption and utilize properties of a specific learning algorithm or predictor class, possibly yielding better performance.

## 7. Conclusions

This paper considered the problem of predicting new feature utility without re-training the original learner. A solution was proposed based on a consistent testing procedure, derived by establishing a function-space relationship between loss gradient and a maximizing transform of the new features. The approach is general, supporting many common learning tasks and loss functions for which the problem is reduced to squared-error regression. This can be performed for just the new features in isolation or in conjuction with existing features. The resulting algorithm allows easy parallelization, making it appropriate for large-scale domains. Empirical evaluation demonstrated the accuracy of the approach on several learning tasks.

**Acknowledgements:** The authors thank Tom Finley for help with ranking experiments and anonymous reviewers for helpful feedback. This work was done while the first author visited Microsoft Research.